\pdfoutput=1

\documentclass[11pt]{article}

\usepackage[final]{acl}

\usepackage{times}
\usepackage{latexsym}
\usepackage{booktabs}

\usepackage[T1]{fontenc}

\usepackage[utf8]{inputenc}

\usepackage{microtype}

\usepackage{inconsolata}

\usepackage{graphicx}
\usepackage{xspace}
\usepackage{multirow}
\usepackage{tabularx}
\usepackage{booktabs}
\usepackage{tcolorbox} 
\usepackage{makecell} 
\usepackage{balance}
\newcommand{\benchmarkname}{\textsc{REPRO-Bench}\xspace}
\newcommand{\agentname}{\textsc{REPRO-Agent}\xspace}
\definecolor{customyellow}{HTML}{6C8EBF}
\definecolor{customyellow1}{HTML}{DAE8FC}
\newcommand{\minihead}[1]{{\vspace{.5em}\noindent\textbf{#1} }}

\title{\benchmarkname: Can Agentic AI Systems Assess the Reproducibility of Social Science Research?}

\author{Chuxuan Hu$^1$, Liyun Zhang$^2$, Yeji Lim$^1$, Aum Wadhwani$^1$, Austin Peters$^3$, Daniel Kang$^1$ \\
$^1$University of Illinois Urbana-Champaign, $^2$Shanghai Jiao Tong University, $^3$University of Chicago \\
\texttt{\{chuxuan3, yejilim2, aumw2, ddkang\}@illinois.edu}, \\ \texttt{zhang\_ly@sjtu.edu.cn}, \texttt{austinpeters@uchicago.edu}}

\begin{document}
\maketitle
\begin{abstract}
Assessing the reproducibility of social science papers is essential for promoting rigor in research processes, but manual assessment is costly. With recent advances in agentic AI systems (i.e., AI agents), we seek to evaluate their capability to automate this process. However, existing benchmarks for reproducing research papers (1) focus solely on reproducing results using provided code and data without assessing their consistency with the paper, (2) oversimplify real-world scenarios, and (3) lack necessary diversity in data formats and programming languages. To address these issues, we introduce \benchmarkname, a collection of 112 task instances, each representing a social science paper with a publicly available reproduction report. The agents are tasked with assessing the reproducibility of the paper based on the original paper PDF and the corresponding reproduction package. \benchmarkname features end-to-end evaluation tasks on the reproducibility of social science papers with complexity comparable to real-world assessments. We evaluate three representative AI agents on \benchmarkname, with the best-performing agent achieving an accuracy of only 21.4\%. 
Building on our empirical analysis, we develop \agentname, which improves the highest accuracy achieved by existing agents by 71\%.
We conclude that more advanced AI agents should be developed to automate real-world reproducibility assessment.
\benchmarkname is publicly available at \url{https://github.com/uiuc-kang-lab/REPRO-Bench}.
\end{abstract}
\section{Introduction}
\label{sec:intro}
\begin{figure}[t!]
    \graphicspath{{figures/}}
    \centering
    \includegraphics[width=\columnwidth]{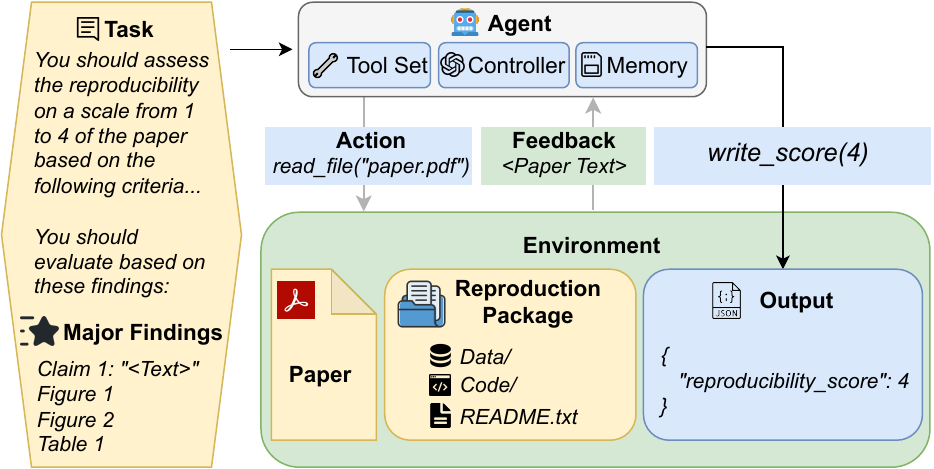}
    \caption{Overview of each \benchmarkname task.}
    \label{fig:workflow}
\end{figure}
To validate social science research findings, domain experts have been seeking systematic methods to assess their reproducibility, from \emph{The Reproducibility Project: Psychology}, which began over a decade ago \cite{OpenScienceCollaboration2012, reproducibility_project2015}, to the recent mass reproduction of economics and political science papers \cite{RePEc:zbw:i4rdps:107}. However, manually assessing the reproducibility of social science papers is costly and time-consuming. For example, 347 social scientists were involved in reproducing 110 papers in the mass reproduction of economics and political science papers \cite{RePEc:zbw:i4rdps:107}, and it took more than five years for \emph{The Reproducibility Project: Psychology} to complete the reproduction of 100 studies \cite{project_wiki}. 

As large language models (LLMs) advance, agentic AI systems (i.e., AI agents) have demonstrated impressive abilities in solving a variety of complex tasks \cite{AutoGPT, siegel2024corebenchfosteringcredibilitypublished, yang2024sweagent}. This opens up new opportunities to 
automate the process of assessing the reproducibility of social science research.
In this paper, we investigate the capability of AI agents in assessing the \emph{computational reproducibility} of social science papers, which evaluates the consistency of the reproduced results with the major findings in the original paper using the originally collected data (Section \ref{sec:bg}, \citealp{RePEc:zbw:i4rdps:107}).

Existing benchmarks with tasks relevant to paper reproduction \cite{siegel2024corebenchfosteringcredibilitypublished, tian2024scicoderesearchcodingbenchmark} have three limitations: (1) they assume all papers are fully reproducible, whereas reproducibility assessment for social science papers \cite{RePEc:zbw:i4rdps:107, OpenScienceCollaboration2012} requires assessing the validity of findings by checking the consistency between reported results and those reproduced from the provided code and data; (2) they provide agents with curated and pre-extracted contexts, while assessing reproducibility for social science papers requires extracting and applying information from original paper PDFs and reproduction packages without prior structuring; and (3) they contain tasks based on a single programming language and/or data format, whereas social science papers often involve multiple languages and data formats, requiring integrated cross-domain knowledge for assessing reproducibility. We address these limitations with further details in Section \ref{subsec:limi}.

To overcome these limitations, we introduce \benchmarkname, consisting of 112 task instances, each representing a social science paper with a public reproduction report (Section \ref{sec:benchmark}). As we illustrate in Figure \ref{fig:workflow}, for each task, the agent is provided with (1) the paper PDF, (2) the reproduction package containing data, code, and documentation, and (3) a list of major findings, and is tasked with generating a reproducibility score on a scale from 1 (least reproducible) to 4 (fully reproducible), following standard reproducibility assessment processes in social science \cite{RePEc:zbw:i4rdps:107}.


\benchmarkname demonstrates three distinct features. 
\emph{Task-wise,} \benchmarkname evaluates agents' critical reasoning capability through paper reproducibility assessment, which involves not only reproducing results but also verifying consistency between the paper and the reproduction package.
\emph{Context-wise,} \benchmarkname provides the full paper PDF and reproduction package for agents to conduct end-to-end reproducibility assessment, simulating real-world scenarios. 
\emph{Complexity-wise,} agents interact with the original reproduction package as the environment in \benchmarkname tasks, ensuring comparable complexity in terms of data and code variety.

We evaluate three representative agents, AutoGPT \cite{AutoGPT}, CORE-Agent \cite{siegel2024corebenchfosteringcredibilitypublished}, and SWE-Agent \cite{yang2024sweagent}, on \benchmarkname. CORE-Agent achieves the highest accuracy of 21.4\%, which is even lower than the expected 25\% accuracy of random guessing among four scores, highlighting the need to build more effective agents for automating social science research reproducibility assessment. We use our findings to develop \agentname, which achieves an accuracy of 36.6\%, a 71\% relative increase in accuracy compared to CORE-Agent.
We detail the experimental setup in Section \ref{sec:models} and analyze results in Section \ref{sec:results}.

\section{Background and Literature Review}
\label{sec:bg}




Since the onset of the reproducibility and replicability crisis (i.e., ``replication crisis'' \cite{Davey2022}), the social science community has placed increasing emphasis on the validity of research results \cite{camerer2018evaluating}. The evaluation of research validity involves the assessment of two major dimensions: \emph{reproducibility} and \emph{replicability} \cite{NAP25303}. Reproducibility refers to the ability to obtain consistent results using the same data and methods as the original study, while replicability refers to achieving robust results using new data but the same methods \cite{RePEc:zbw:i4rdps:107}. 

We study the capabilities of AI agents in assessing the \emph{computational reproducibility} of social science papers. We focus on reproducibility using the original raw data, without re-executing the data collection process or introducing new data. Following formal definitions and guidelines \cite{NAP25303, nsf2023}, agents are required to assess the validity of research findings by checking the consistency between reproduced and reported results using provided data processing and analysis scripts. We review the importance of computational reproducibility in social science research in Section \ref{subsec:cr_importance} and highlight the limitations of existing benchmarks in evaluating AI agents' ability to assess reproducibility in Section \ref{subsec:limi}.

\subsection{Importance of Reproducibility}
\label{subsec:cr_importance}
Assessing reproducibility from both the perspective of computational results and code validity is essential for the development of social science from the following three dimensions.

First, reproducibility offers a more direct and reliable evaluation of social science research by enforcing strict standards compared to replicability \cite{RePEc:zbw:i4rdps:107}. The Social Science Reproduction Platform (SSRP) provides a breakdown of reproducibility levels, focusing on data and code availability, ranging from level 1, where data and code are missing, to level 10, where the study is fully computationally reproducible \cite{ssrp}. 
With fully available data and code, \citet{RePEc:zbw:i4rdps:107} and \citet{OpenScienceCollaboration2012} further standardize the process as a two-phase procedure: first, the provided data and code are examined, and then the major findings are reproduced and compared.

Second, investigating code validity is important because irreproducibility can occur due to coding errors. For example, in the reproduction report \cite{chen2024many} for \citet{NBERw24826}, the authors mention a coding error that \textit{``assigns a value of zero for the variable \emph{of color} to both individuals identified as \emph{white} and \emph{other} in the raw data.''} After fixing this error, the major findings changed significantly. 

Third, even though reproducibility seems to be a basic, minimal requirement for social science research findings, existing reproduction results reveal insufficiencies in ensuring such guarantees. As of 06/27/2024, less than 40\% of the papers reproduced on SSRP \cite{ssrp} are considered fully reproducible at level 10. In a recent large-scale reproduction of economic and political science papers, 25\% of the reproduced papers contained coding errors, even when excluding minor issues like missing packages or misconfigured file paths, with a considerable portion of studies exhibiting multiple errors \cite{RePEc:zbw:i4rdps:107}.


\subsection{Existing AI Agent Benchmark Limitations}
\label{subsec:limi}

AI agent benchmarks \cite{yao2023webshopscalablerealworldweb, jimenez2024swebenchlanguagemodelsresolve, tian2024scicoderesearchcodingbenchmark, siegel2024corebenchfosteringcredibilitypublished, hu2025leap} evaluate the reasoning capabilities of AI agents through complex tasks \cite{sun2024surveyreasoningfoundationmodels}, driving their continuous improvement. However, existing paper reproduction benchmarks have three key limitations in evaluating AI agents' ability to assess 
reproducibility.

First, existing benchmarks assume all papers are fully reproducible, whereas reproducibility assessment requires evaluating both the validity of major findings and the consistency of provided code and data \cite{RePEc:zbw:i4rdps:107, OpenScienceCollaboration2012}. 
SciCode tasks \cite{tian2024scicoderesearchcodingbenchmark} are coding problems based on the major findings, assuming they are valid, and CORE-Bench tasks \cite{siegel2024corebenchfosteringcredibilitypublished} consider the execution results of the provided code on the provided data as ground truth, assuming they are consistent.

Second, existing benchmark contexts are overly pre-processed and curated. Each SciCode task \cite{tian2024scicoderesearchcodingbenchmark} is a highly condensed problem derived from paper findings, and each CORE-Bench task \cite{siegel2024corebenchfosteringcredibilitypublished} represents a pre-extracted, concrete step from the paper's reproduction process. However, in actual social science reproducibility assessment, reproducers are not given predefined steps; instead, they must independently analyze the paper PDFs and reproduction packages, extract relevant information, and formulate a plan to inspect the code and data for potential inconsistencies \cite{RePEc:zbw:i4rdps:107, OpenScienceCollaboration2012}.

Third, existing benchmarks contain tasks based on a single programming language and/or data format. 
SciCode \cite{tian2024scicoderesearchcodingbenchmark} tasks require models to generate Python code, and each code repository in CORE-Bench \cite{siegel2024corebenchfosteringcredibilitypublished} contains code in a single language, either R or Python. 
However, each social science paper typically involves diverse programming languages and multiple data formats, requiring integrated knowledge across various domains to effectively assess reproducibility. For example, to reproduce the major findings in \citet{paper2}, one must first execute a Stata script to analyze \texttt{.dta} data for Study 1, and then run an R project to analyze \texttt{.csv} data for Study 2.
\section{\benchmarkname }
\label{sec:benchmark}
We propose \benchmarkname, a benchmark for evaluating AI agents' capability to assess the reproducibility of social science papers. 
We describe our data collection process in detail in Section \ref{subsec:data} and explain the methodology for determining ground-truth reproducibility scores in Section \ref{subsec:repro_score}. We present a detailed statistical analysis of \benchmarkname in Section \ref{subsec:data_stat}. 
Based on these, we formally define the \benchmarkname tasks and 
describe their key features 
in Section \ref{subsec:task}.

\subsection{Data Collection}
\label{subsec:data}
We collect 112 papers across 4 sources, with each paper representing a task instance. Source 1 serves as our primary source, where a significant portion of papers are reported as largely or fully reproducible. To effectively assess AI agents' ability to identify sources of inconsistencies, we further incorporate Sources 2–4, which contain papers with crucial reproducibility issues.
We apply the following universal criteria $\mathcal{C}$ for all sources: 
\begin{itemize}
    \item To ensure the specificity of \benchmarkname tasks, each paper must be ($\mathcal{C}_1$) published in the social science field.
    \item To ensure public accessibility, each paper must ($\mathcal{C}_2$) have a valid DOI and ($\mathcal{C}_3$) include a publicly available reproduction package.
    \item To ensure that the reproducibility of the paper is verifiable, each paper must ($\mathcal{C}_4$) have a credible public reproduction report that thoroughly investigates the reproduced results, is authored by social science experts, and is influential and highly regarded.
    \item To prevent the benchmark from being overly time-consuming, we require ($\mathcal{C}_5$) that either the reproduction package's README file or the reproduction report explicitly state that the paper's reproduction time is less than 2 hours.
\end{itemize}

To ensure that \benchmarkname includes both recent social science papers and up-to-date reproduction efforts, we account for different data ranges based on the nature of the sources. We select based on publication dates of the original papers or reproduction reports to ensure a balanced and representative sample.
We present the source-specific criteria, including data ranges, as follows. 

\minihead{Source 1: Mass Reproducibility and Replicability of Social Science Papers.} To enhance the understanding of research reliability, researchers reproduced and replicated the major findings from 110 papers in leading economic and political science journals \cite{RePEc:zbw:i4rdps:107}. Since this mass reproduction effort already focuses on recent and influential papers, we do not impose additional data range restrictions other than $\mathcal{C}$, resulting in the selection of 92 papers from the original 110. This mass reproduction has been cited 30 times as of March 30, 2025, within just one year of publication. For reference, the top 1\% most-cited economics paper, published in 1991, has 354 citations as of the same date \cite{RePEcTopCitations}.

\minihead{Source 2: Institute For Replication (I4R)'s Discussion Paper Series.\footnote{\href{https://i4replication.org/discussion_paper.html}{https://i4replication.org/discussion\_paper.html}}} I4R facilitates reproductions and replications to enhance the credibility of research findings. In addition to $\mathcal{C}$, we apply the following criteria to I4R's discussion paper series: (1) since I4R began actively and systematically updating in 2024, we select papers with reproduction results published between 01/01/2024 and 09/30/2024, and (2) the reproduction results identify errors and/or issues in the data and/or code. This resulted in the selection of 11 papers.

\minihead{Source 3: Retraction Watch Database.\footnote{\href{https://retractionwatch.com/}{https://retractionwatch.com/}}} The Retraction Watch database consists of retracted papers. Using its search engine, we apply the following filters: (1) \emph{Subject(s):} Social Sciences (SOC) ($\mathcal{C}_1$); (2) \emph{Reason(s) for Retraction:} Error in Data OR Error in Analyses OR Error in Results and/or Conclusions OR Error in Materials (General) OR Error in Methods; (3) \emph{Original Paper Date:} since retraction processes are lengthy and complex, we selected papers with original publication dates from a broader range (01/01/2019 to 01/01/2024) to ensure sufficient verification ($\mathcal{C}_4$). We did not select based on the publication dates of reproduction reports, as long retraction timelines could result in selecting very old papers despite recent reproduction efforts.
We further applied the remaining $\mathcal{C}$ criteria and excluded PDFs containing a \emph{Retracted} watermark, resulting in a final selection of 7 papers.

\minihead{Source 4: Twitter/X.} 
Important reproduction efforts also occur outside the academia and formal publications. For example, significant reproduction efforts are actively discussed on social media like Twitter/X. 
We collect 2 social science research papers that meet $\mathcal{C}$ and have been identified as having reproducibility issues in Tweets posted between 01/01/2024 and 06/30/2024. Given the frequent updates on Twitter/X, we apply a shorter time range. 
In this context, we define reproduction reports as the full reports and code linked within the Tweets, rather than the Tweets themselves.
These two Tweets have received 61.5K and 18.7K views as of March 30, 2025, respectively, indicating strong public engagement.

\vspace{.5em}

We annotate the major findings of each paper as a list of all the items (i.e., \textit{figures}, \textit{tables}, and \textit{text claims}) reproduced in the corresponding reproduction report ($\mathcal{C}_4$). For example, in the reproduction report \cite{paper1-report} for
\citet{paper1} (Paper 1), the reproducers reproduced Table 5 of the original paper in Table 1 of the report and reproduced Figure 6 of the original paper in Figure 1 of the report. Thus, we annotate the major findings of Paper 1 as a list 
\texttt{["Table 5", "Figure 6"]}.
Text claims refer to experimental results reported as in-line texts rather than in figures or tables. They are extracted exactly as they appear in the original paper. For example, in the reproduction report \cite{RePEc:zbw:i4rdps:43} of Paper 40 \cite{RePEc:oup:restud:v:89:y:2022:i:5:p:2293-2328.}, the reproducers reproduced the following textual results in the reproduced paper: \textit{``In line with the pronounced visual differences, the distributions of attention spans differ significantly across treatments (Kolmogorov–Smirnov test, p < 0.001 for all pairwise treatment comparisons).''}

Given the minimal subjectivity involved in identifying major findings, we adopt a consensus-based annotation process. The lead author first manually extracts the reproduced findings from each paper's official reproduction report. These findings are then cross-verified by the rest of the five-member team, which includes a legal expert with deep familiarity with the structure and language of social science documentation. Full agreement is reached before finalizing the annotations.

\subsection{Reproducibility Score}\label{subsec:repro_score}


The SSRP reproducibility metric \cite{ssrp} primarily assesses data and code availability, with levels 1–8 focusing on whether data and/or code are provided, while only levels 9 and 10 address actual reproducibility. To maintain task complexity, we apply twice as fine-grained metrics following the social science reproducibility assessment standards by \citet{RePEc:zbw:i4rdps:107}, ensuring it accurately captures the nuanced nature of reproducibility in social science research.
We annotate the reproducibility score of the papers based on their public reproduction reports, using a scale from 1 to 4, where different scores account for varying levels of consistency between the paper PDF and the corresponding data and code in the reproduction package. 
The scoring criteria are
defined 
as follows:
\begin{itemize}
    \item 1: major findings are irreproducible.
    \item 2: there are minor inconsistencies and/or errors in the provided code.
    \item 3: there are minor issues at the display and reporting level, e.g., rounding errors.
    \item 4: major findings are fully reproducible.
\end{itemize}

Scores 1 and 4 reflect binary reproducibility outcomes, while scores 2 and 3 capture more nuanced issues. A score of 2 indicates identifiable inconsistencies in the code that do not alter the paper’s major findings. For example, the reproduction report \cite{RePEc:zbw:i4rdps:97} for a score-2 paper \cite{doi:10.1086/714924} notes: \textit{``We do not find any major coding errors. One minor point that we find is that there is some inconsistency in how NA values are coded for the gender variable.''} While this coding issue affects the data structure, it does not compromise the core findings.
A score of 3 is assigned when the analysis and calculations are correct, but minor reporting discrepancies are observed. For instance, in the reproduction report \cite{RePEc:zbw:i4rdps:29} for a score-3 paper \cite{RePEc:oup:econjl:v:132:y:2022:i:647:p:2392-2411.}, it is stated: \textit{``When we perform the calculation to more precision, it is revealed as 3.848456. This suggests that an initial calculation rounded to two decimal places (3.85), followed by another rounding to one decimal place, produced 3.9.''}

Table \ref{tab:stat} shows the reproducibility score distribution, with balanced counts of papers scoring 1–2 (indicating recognizable reproducibility issues) and 3–4 (largely or fully reproducible) to ensure a fair evaluation, 
further demonstrating the effectiveness of 
our data collection criteria.

All reproduction efforts in social science go through the standard reproducibility assessment pipeline \cite{RePEc:zbw:i4rdps:107}, where the reproducibility assessment can be deterministically classified into the four scores we defined. Given this, the manual annotations of ground-truth reproducibility is an objective task with the design that ensures consistency across different sources and reproduction efforts.
The manual annotation process follows a rigorous and consensus-driven procedure. Specifically, the lead author manually labels each paper’s reproducibility score based on the official reproduction report. These labels are then cross-verified by the rest of the team of 5, including a legal expert who has deep familiarity with the structure and language of social science documentation. Full agreement is reached through a structured, consensus-based review process, using the same scoring criteria across all sources.
\begin{table}[ht]
\centering
\begin{tabularx}{\columnwidth}{
    >{\centering\arraybackslash}X
    >{\centering\arraybackslash}X
    >{\centering\arraybackslash}X
    >{\centering\arraybackslash}X
}
\specialrule{1.5pt}{0.5pt}{0.5pt}
\multicolumn{4}{c}{\textbf{\small Reproducibility Score Distribution}} \\ 
\specialrule{1.5pt}{0.5pt}{0.5pt}
\small Score 1 & \small Score 2 & \small Score 3 & \small Score 4 \\
\hline
\small 20 & \small 36 & \small 8 & \small 48 \\
\specialrule{1pt}{0pt}{0pt}
\multicolumn{2}{c}{\small Score 1+ Score 2} & \multicolumn{2}{c}{\small Score 3 + Score 4} \\
\hline
\multicolumn{2}{c}{\small 56} & \multicolumn{2}{c}{\small 56} \\
\end{tabularx}


\begin{tabularx}{\columnwidth}{
    >{\centering\arraybackslash}X
    >{\centering\arraybackslash}p{0.01\textwidth}
    >{\centering\arraybackslash}X
    >{\centering\arraybackslash}X
    >{\centering\arraybackslash}X
    | 
    >{\centering\arraybackslash}p{0.08\textwidth} 
}
\specialrule{1.5pt}{0pt}{0pt}
\multicolumn{6}{c}{\textbf{\small Programming Language}} \\  
\specialrule{1.5pt}{0pt}{0pt}
\multicolumn{5}{c|}{\small Single Language} & \multirow{2}{*}{\makecell{\small Multiple \\ \small Languages}} \\  
\cline{1-5}
\small Stata & \small R & \small MATLAB & \small Julia & \small Python & \\  
\cline{1-6}
\small 63 & \small 25 & \small 2 & \small 1 & \small 1 & \small 15 \\  
\end{tabularx}

\begin{tabularx}{\columnwidth}{
    >{\centering\arraybackslash}X
    >{\centering\arraybackslash}X
    >{\centering\arraybackslash}X
    >{\centering\arraybackslash}X
    >{\centering\arraybackslash}X
    | 
    >{\centering\arraybackslash}p{0.06\textwidth} 
}
\specialrule{1.5pt}{0pt}{0pt}
\multicolumn{6}{c}{\textbf{\small Data Formats}} \\  
\specialrule{1.5pt}{0pt}{0pt}
\multicolumn{5}{c|}{\small Single Format} & \multirow{2}{*}{\makecell{\small Multiple \\ \small Formats}} \\  
\cline{1-5}
\small .dta & \small .csv & \small .rda(ta) & \small .xls(x) & \small .sav & \\  
\cline{1-6}
\small 34 & \small 11 & \small 10 & \small 5 & \small 1 & \small 51 \\  
\specialrule{1.5pt}{0pt}{0pt}
\end{tabularx}

\caption{Statistics of \benchmarkname.}
\label{tab:stat}
\end{table}


\subsection{Data Statistics and Analysis} \label{subsec:data_stat}

\benchmarkname\ includes papers that average 29 pages in length. The corresponding reproduction packages average 4.2 GB in size and contain 142 files, spanning various programming languages and diverse social science data formats (Table \ref{tab:stat}).
On average, each paper has 5 major findings, with the actual number ranging from 1 to 19 and a standard deviation of 4. 

To validate that paper reproducibility is unaffected by irrelevant factors, we compute Spearman correlation coefficients $\rho$ \cite{spearman1904} between different paper features and ground-truth reproducibility scores. We encode single data format or programming language as 0 and multiple as 1, and apply label encoding to represent different data formats and programming languages as numerical values. 
The results in Table \ref{tab:language_data_comparison} show that all factors have
$|\rho| < 0.1$, indicating no meaningful correlations with ground-truth reproducibility scores. 
This confirms that these factors do not impact reproducibility, 
supporting the rigor of our data collection process 
and benchmark design.

\begin{table}[h]
    \centering
    \small
    \begin{tabular}{l r}
\specialrule{1.5pt}{1pt}{2pt}
        \# pages & -0.064 \\
        \# major findings & -0.034 \\
        Reproduction package size (MB) & 0.023 \\
        \# files in reproduction package & 0.0084 \\
        Single vs. multiple programming languages & -0.0035 \\
        Single vs. multiple data formats & 0.057 \\
        Different programming languages & 0.044 \\
        Different data formats &  -0.011 \\
\specialrule{1.5pt}{0.0pt}{0pt}
    \end{tabular}
    \caption{The correlation coefficients $\rho$ between different paper features and ground-truth reproducibility scores.}
    \label{tab:language_data_comparison}
\end{table}

\subsection{Task Formulation} \label{subsec:task}

As we illustrate in Figure \ref{fig:workflow}, following the standard process for assessing the reproducibility of social science papers \cite{RePEc:zbw:i4rdps:107}, we define each task instance in \benchmarkname as follows: an agent is provided with (1) a social science paper (in PDF format), (2) the corresponding reproduction package, including data, code, and documentations, and (3) a list of major findings. The agent is tasked with generating a reproducibility score based on the scoring criteria in Section \ref{subsec:repro_score}. 
To ensure consistent formatting for large-scale data collection in real-world applications, the agent is instructed to generate a valid output file, \texttt{reproducibility\_score.json}, containing a single entry named \texttt{reproducibility\_score}, with the score stored as an integer value. This file must be placed in the root folder where the agent starts executing.
\benchmarkname contains tasks demonstrating the following three distinct features.

\minihead{Real-world tasks impacting actual social science research.} 
\benchmarkname\ tasks mirror recent social science reproducibility assessments \cite{RePEc:zbw:i4rdps:107, OpenScienceCollaboration2012}, requiring agents to assess the reproducibility of real-world papers end-to-end.
According to a legal expert, \benchmarkname captures representative patterns of social science papers and inspires efficient reproducibility assessment tools, promoting better code and data management for social science researchers.

\minihead{Complex tasks involving long and diverse contexts.} 
To complete tasks in \benchmarkname, agents must extract essential information from long paper texts and large volumes of data, while handling multiple data formats and programming languages, to assess paper reproducibility.


\minihead{Evaluation tasks requiring critical reasoning.} \benchmarkname offers insights into AI agents' application for evaluation tasks similar to assessing research reproducibility. 
To generate accurate reproducibility scores, agents are required to demonstrate a wide range of reasoning skills:
logical reasoning to interpret papers and code, mathematical reasoning to modify and run code, visual reasoning to examine data points, and causal reasoning to infer scientific insights from results. Beyond these reasoning skills widely studied in existing benchmarks \cite{sun2024surveyreasoningfoundationmodels}, the nature of \benchmarkname uniquely demands strong \textit{critical reasoning}. This foundational skill, which cannot be reduced to simpler forms of reasoning, is essential for identifying and evaluating discrepancies between original and reproduced results.



\section{Experiment Setup}
\label{sec:models}
We introduce the agent environment, actions, and feedback in Section \ref{subsec:enc_act_feedback}, selected agents in Section \ref{subsec:agents}, and evaluation 
metrics in Section \ref{subsec:metric}.

\subsection{Environment, Actions, and Feedback} \label{subsec:enc_act_feedback}
Each agent starts in a directory containing \texttt{paper.pdf} and a subdirectory \texttt{reproduction\_package/}, with task descriptions in the user prompt that include the major findings to be reproduced. 
All necessary software, including Stata, MATLAB, and LaTeX, is preinstalled in the environment, with version details specified in the task descriptions.
Agents have the freedom to execute any command line operations, install necessary packages, and access all files on the system.
They receive feedback from the environment through both standard output and standard error streams of the executed commands. 
To maintain objectivity, we ensure that agents operate without access to underlying data distributions or results from other task instances.

\subsection{Agents} \label{subsec:agents}
We select and adapt the following three agents to perform \benchmarkname tasks.

\minihead{AutoGPT \cite{AutoGPT}.} AutoGPT is a generalized agent designed for a wide range of tasks, with capabilities that include making long-term plans, selecting and using tools, and reflecting on past actions.
We select AutoGPT to investigate the capability of general-purpose agents in solving \benchmarkname tasks.

\minihead{CORE-Agent \cite{siegel2024corebenchfosteringcredibilitypublished}.} CORE-Agent is capable of completing subtasks within scientific papers. We select the version of CORE-Agent specifically adapted for hard tasks in CORE-Bench, including its vision language model (VLM) tool, to investigate the complexity of assessing the reproducibility of a social science paper from scratch in \benchmarkname tasks, in comparison to reproduction using predefined, concrete commands.

\minihead{SWE-Agent \cite{yang2024sweagent}.} SWE-Agent is a software engineering agent capable of resolving real-world GitHub issues. 
We select SWE-Agent to investigate how its Agent-Computer Interface (ACI) supports the execution and debugging of reproduction packages given social science papers in \benchmarkname tasks.

\vspace{.5em}
All three agents use \texttt{gpt-4o-2024-05-13} \cite{gpt4o}. Following the original settings of all three agents, we terminate the agents if they incur API costs of over \$4 per task.

\subsection{Metrics} \label{subsec:metric}

For performance evaluation, we use accuracy as the primary metric, measuring whether the generated reproducibility score matches the ground truth.
We examine applicability rates to verify whether the agent generates valid outputs following the instructions in Section \ref{subsec:task}. Validity is evaluated in two dimensions: the output file (1) must follow the correct format and naming convention \href{https://anonymous.4open.science/r/REPRO-Bench-Code-2E98/SWE-agent/evaluation.py}{[code]}, and (2) must be placed in the root directory where the agent starts executing \href{https://anonymous.4open.science/r/REPRO-Bench-Code-2E98/SWE-agent/run_reprobench.sh}{[code]}.
We report both the original and adjusted accuracy and applicability rates, with the adjusted versions accounting for cases where agents generate valid output files outside the designated directory (i.e., satisfies only (1) but not (2)).
For cost analysis, we report the average API costs for all requests made by each agent for each 
task.

\section{Experiment Results}
\label{sec:results}
We present the quantitative results in Section \ref{subsec:overall_res}, analyze agent reasoning traces through case studies in Section \ref{subsec:case_studies}, and validate our findings by developing \agentname 
with significant performance improvements in Section \ref{subsec:repro-agent}.






\subsection{Quantitative Analysis}\label{subsec:overall_res}
We report the overall success rates, applicability rates, and costs in Table \ref{tab:model_performance}. 
CORE-Agent achieves the highest accuracy at 21.4\% among the three agents, which is still lower than the expected 25\% accuracy of random guessing among four options without prior knowledge of the underlying data distributions or the results of other task instances. Although CORE-Agent is designed for paper reproduction tasks, its accuracy is only slightly higher (by less than 1\%, representing just one additional correct task) than the general-purpose AutoGPT. SWE-Agent exhibits the lowest performance, with only 10.7\% accuracy even after adjustments, indicating that simple ACI actions are insufficient for handling the complex tasks in \benchmarkname. 

\begin{table}[h!]
\centering
\resizebox{\columnwidth}{!}{%
\begin{tabular}{lccc}
\toprule
Agent & \% Accuracy & \% Applicability & Cost (\$) \\
\midrule
AutoGPT        & 20.5 & \textbf{60.7} & 2.03  \\
CORE-Agent     & \textbf{21.4} & 46.4 & 2.00  \\
SWE-Agent      & 1.8 (10.7) & 1.8 (19.6) & 1.20 \\
\bottomrule
\end{tabular}
}
\caption{Performance and costs of different agents on \benchmarkname. Adjusted values following Section \ref{subsec:metric} are reported in brackets if differ from original values.}
\label{tab:model_performance}
\end{table}

We now analyze the performance in detail. We use the adjusted values to derive more statistically significant conclusions.

\begin{figure*}[t!]
    \graphicspath{{figures/}}
    \centering
    \includegraphics[width=\textwidth]{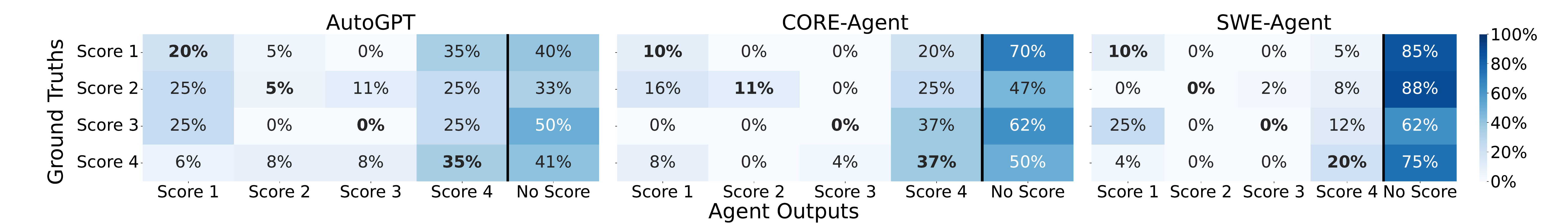}
    \caption{Agent outputs across different reproducibility scores. Diagonal values (bold) represent accuracy. \emph{No Score} on the prediction axis refers to cases where AI agents did not generate valid outputs.
}
    \label{fig:score}%
\end{figure*}

\minihead{Agents are better at identifying reproducible papers.} We present the distribution of generated reproducibility scores in Figure \ref{fig:score}. We can see that all three agents perform significantly better on papers with a reproducibility score of 4. Furthermore, agents tend to perform better on reproducibility scores of 1 and 4 compared to scores of 2 and 3, suggesting that the agents are inclined to generate binary results rather than thoroughly investigating the sources of inconsistencies.

\begin{figure}[t!]
    \graphicspath{{figures/}}
    \centering
    \includegraphics[width=\columnwidth]{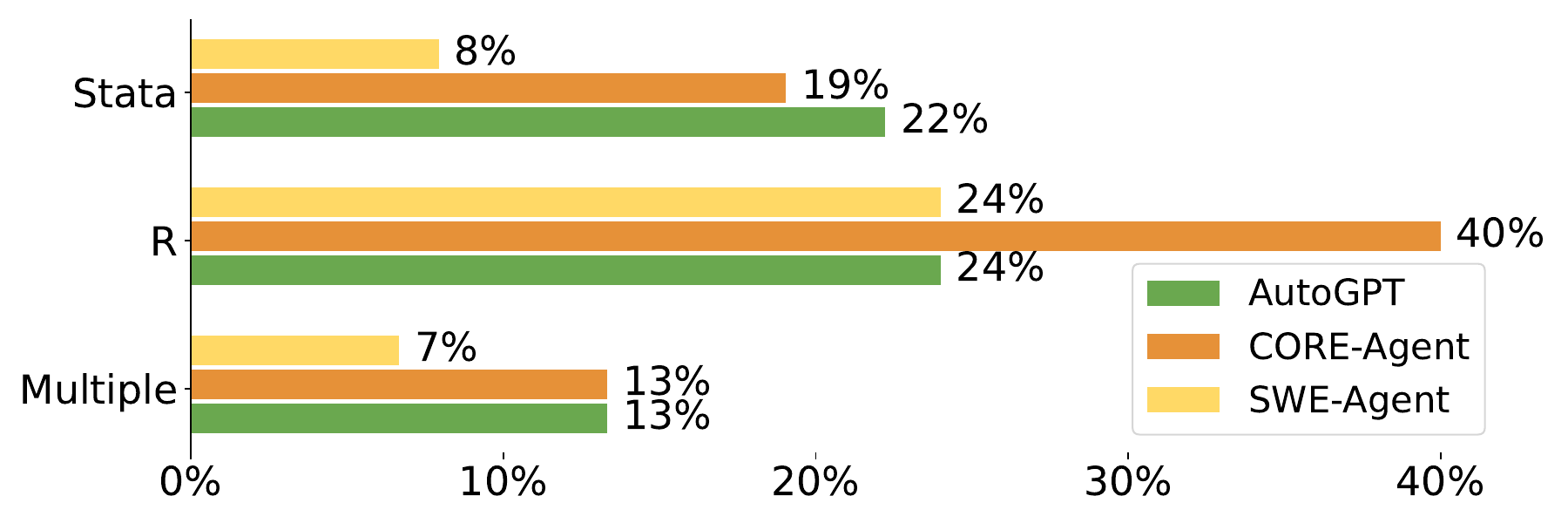}
    \caption{Agent accuracy across different languages.}
    \label{fig:language}%
\end{figure}

\minihead{Agents are better at R than Stata.} We summarize the accuracy distributions across task instances with reproduction packages using the major programming languages in social science research, Stata and R, as well as those using multiple programming languages, in Figure \ref{fig:language}. We can see that all three agents perform significantly better on tasks with R code compared to those with Stata. This is because, unlike Stata, which requires a purchased license, R is more widely used across all domains, and therefore, LLMs are likely to have better knowledge of it.

\minihead{Agents underperform when tasked with handling multiple programming languages.} As we can see from Figure \ref{fig:language}, the accuracy of all three agents drops in task instances with multiple programming languages compared to those with a single programming language. This indicates that LLMs struggle to ensure consistent execution across diverse programming languages.



\minihead{Data in multiple formats does not introduce performance degradation.}
We compare the accuracy of the three agents on task instances where reproduction packages contain source data in either a single or multiple formats. Despite data variety, the agents achieve comparable accuracy: the average accuracy across all three agents is 54\% for tasks with a single data format and 52\% for tasks with multiple data formats. This indicates that LLMs are capable of leveraging data loaders to effectively integrate knowledge from data of diverse formats.

\subsection{Qualitative Analysis}\label{subsec:case_studies}
We analyze agent traces on \benchmarkname to demonstrate agents' ability to reason critically and autonomously perform 
complex assessments.

\begin{figure}[t!]
    \graphicspath{{figures/}}
    \centering
    \includegraphics[width=0.9\columnwidth]{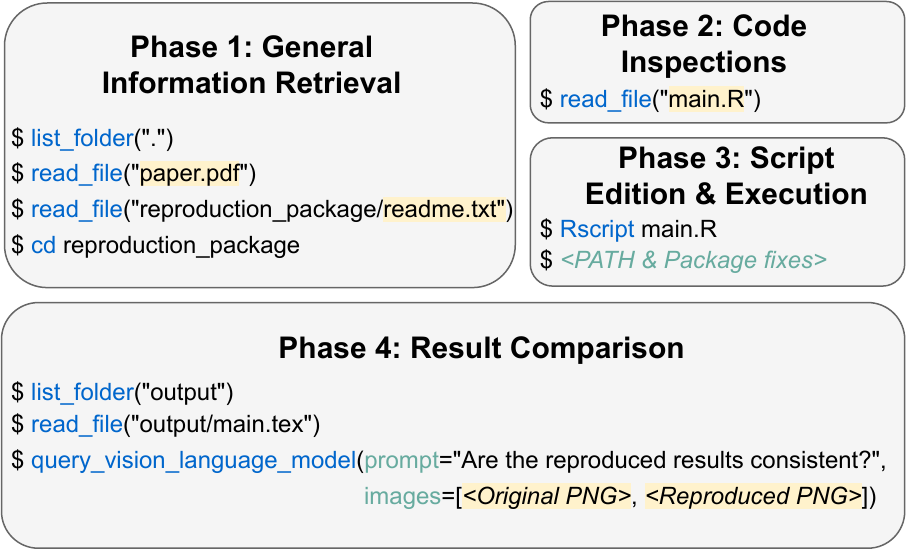}
    \caption{Agent workflow for \benchmarkname tasks, exemplified by the traces of CORE-Agent for Task 4.}
    \label{fig:success}%
\end{figure}

\minihead{Workflow of AI agents.} By analyzing the traces of all successful cases, we outline a general workflow of the agents in Figure \ref{fig:success}. Specifically, the agents start by developing a broad understanding of the task and the environment in Phase 1, where they (1) list all files and directories in the workspace to identify available materials, (2) read the paper, and (3) read the README file. In Phase 2, they inspect the provided code for potential inconsistencies. In Phase 3, they edit and execute scripts, and in Phase 4, they compare the execution results with the original results. This workflow closely aligns with real-world scenarios \cite{RePEc:zbw:i4rdps:107}, demonstrating the effectiveness of \benchmarkname tasks. 

\begin{figure}[t!]
    \graphicspath{{figures/}}
    \centering
    \includegraphics[width=0.9\columnwidth]{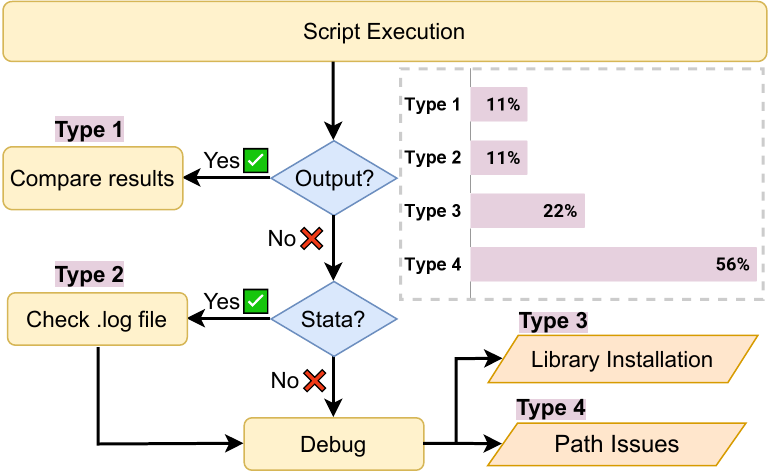}
    \caption{Occurrences and distributions of 4 types of failure sources in reproducing social science papers.}
    \label{fig:command workflow}%
\end{figure}

\minihead{Why do AI agents fail to reproduce and analyze results?} By analyzing all the traces that misclassify score 4 tasks as score 1 across the 3 agents, we summarize the general workflow of executing scripts and comparing results (i.e., Phases 3 and 4 in Figure \ref{fig:success}), categorize the sources of failure to reproduce results in social science papers into 4 types, and illustrate their distributions in Figure \ref{fig:command workflow}.

Type 1 failures occur when result comparison is incorrect. For example, CORE-Agent wrote an erroneous Python script for comparison in Task 50, falsely classifying consistent results as unmatched. 
Type 2 failures occur when agents see no terminal output for Stata scripts because error messages are stored in log files rather than printed in the terminal, leading them to conclude that the script does not produce consistent results, as SWE-Agent did for Task 76.
Type 3 failures occur when agents cannot correctly install libraries. 
Type 4 failures occur when agents fail to locate files due to incorrect directory placement. For example, in Task 62, the reproduction package contains all the required data, but \texttt{BGLM\_Data.dta} and \texttt{DuvEq-12-24-50.txt} are not in the code execution directory, resulting in file missing errors. AutoGPT and CORE-Agent incorrectly concluded there are missing data without searching the package. 

As we illustrate in Figure \ref{fig:command workflow}, Type 4 failures occur most frequently. This is because the organization of reproduction packages is not as straightforward as in traditional code repositories like SWE-Bench \cite{jimenez2024swebenchlanguagemodelsresolve} and CORE-Bench \cite{siegel2024corebenchfosteringcredibilitypublished}. Thus, the agents must infer the directory layout by inspecting package structures and README files. Our results indicate that existing agents lack proficiency in effectively navigating and interpreting these complex directory structures. 

\minihead{Why do AI agents fail to recognize inconsistencies?} 
We inspect all traces where score 1 tasks are misclassified as score 4 across the three agents and identify two primary reasons for overlooking major inconsistencies. First, the agents do not strictly follow the workflow outlined in Figure \ref{fig:success}. 
Notably, in less than half (42\%) of cases, agents incorporate both Phase 2 (code inspection) and Phase 4 (result comparison) in their workflows, despite their crucial role in detecting inconsistencies.
Second, in Phase 2, agents often read entire code files instead of focusing on relevant sections, making error identification difficult due 
to 
long code context.











\subsection{\agentname}
\label{subsec:repro-agent}
We apply the empirical analysis in Sections \ref{subsec:overall_res} and \ref{subsec:case_studies} to build \agentname. 
\agentname addresses common failure patterns in existing agents through three key strategies:  (1) following a structured template built upon successful reproducibility assessment cases to improve \textit{planning}; (2) incorporating a dummy score prediction as a \textit{fallback mechanism}; and (3) using common error sources as few-shot examples to enhance \textit{in-context learning} effectiveness.
Specifically, we adjust CORE-Agent with the following additional instructions:

\begin{tcolorbox}[colback=customyellow1, colframe=customyellow, title=\small\agentname]
\small
\emph{Success Case Template (Figure \ref{fig:success})}\\
- You should follow this general workflow of four phases:…

\emph{Low Applicability (Table \ref{tab:model_performance})}\\
- You should always generate a dummy score in the first step…

\emph{Common Error Sources (Figure \ref{fig:command workflow})}

- If you are using Stata, remember that the error messages are stored in log files rather than displayed directly in the terminal.

- In some cases, the data files are provided but not in the folder as indicated in the README files…

\end{tcolorbox}

\agentname achieves an accuracy of 36.6\%, a relative improvement of 71\% compared to CORE-Agent, which had the highest accuracy (21.4\%) among existing AI agents. With the strategy of first generating a dummy score and then refining it afterward, \agentname achieves an applicability rate of 92.9\%, a relative improvement of 53\% over AutoGPT, which had the highest applicability rate (60.7\%) among existing AI agents. 

\agentname's significantly improved performance validates our two key contributions: (1) we systematically identify deficiencies and specific failure modes in existing AI agents, and (2) we demonstrate concrete and effective directions to address the limitations of existing AI agents.
\section{Conclusion}
\label{sec:conclusion}
We introduce \benchmarkname, a benchmark designed to evaluate the capability of AI agents in assessing the reproducibility of social science papers. We evaluate three representative agents on \benchmarkname, with the highest accuracy reaching only 21.4\%. Building on our empirical findings, we develop \agentname, which achieves a 71\% relative improvement in accuracy, reaching 36.6\%. However, this performance remains insufficient for practical applications, highlighting the need for developing more powerful AI agents with enhanced reasoning capabilities, better contextual understanding, and robust evaluation frameworks. 
\section{Limitations}
\label{sec:limitations}
Our work has the following limitations that could be addressed in future work:
\begin{itemize}
    \item Lack of alternative versions of task instances: While \benchmarkname's current design already presents a robust, challenging, and reasonable evaluation, as evidenced by the observed performance differences across agents, we believe its granularity can be further improved by introducing multiple versions of task instances for the same paper, incorporating intentionally erroneous or corrected code and/or data.
    \item Lack of investigations into more complex scenarios: \benchmarkname follows real-world scenarios, where reproducers have access to the entire paper and the reproduction package.
    Future work can explore the capability of agents to reproduce social science papers in more challenging settings by masking the data points in the experiment results and providing the agents only with raw data.
    
    \item Extension into diverse domains: Beyond social science papers, where large-scale reproduction efforts are already underway, \benchmarkname can be extended to other fields where reproducibility is critical, such as biology \cite{begley2015reproducibility}, to more comprehensively evaluate the ability of agents to reproduce research findings.
    \item Building more advanced agents: Inspired by \benchmarkname, more powerful agents than \agentname can be developed to better accommodate the growing need for automating reproduction processes in social science.
    \item Potential for large-scale automation: The annoation process from reproduction reports can potentially be automated. A promising direction involves a lightweight pipeline that combines OCR, pattern-based extraction, and LLM-based claim identification from reproduction reports.
\end{itemize}
\section*{Acknowledgments}
We thank Yuxuan Zhu, Qiusi Zhan, Lilia Tang, and Tengjun Jin for their feedback and help.

\bibliography{custom}

\begin{thebibliography}{32}
\providecommand{\natexlab}[1]{#1}

\bibitem[{Akhtar and Ye(2023)}]{RePEc:zbw:i4rdps:29}
Ahwaz Akhtar and Hao Ye. 2023.
\newblock \href {https://ideas.repec.org/p/zbw/i4rdps/29.html} {{Reproducibility and Robustness Replicability of Gsottbauer et al. (2022)}}.
\newblock I4R Discussion Paper Series~29, The Institute for Replication (I4R).

\bibitem[{Altmann et~al.(2022)Altmann, Grunewald, and Radbruch}]{RePEc:oup:restud:v:89:y:2022:i:5:p:2293-2328.}
Steffen Altmann, Andreas Grunewald, and Jonas Radbruch. 2022.
\newblock \href {https://ideas.repec.org/a/oup/restud/v89y2022i5p2293-2328..html} {{Interventions and Cognitive Spillovers}}.
\newblock \emph{The Review of Economic Studies}, 89(5):2293--2328.

\bibitem[{Bachler et~al.(2023)Bachler, Erhart, and Holzknecht}]{RePEc:zbw:i4rdps:43}
Sebastian Bachler, Andrea Erhart, and Armando Holzknecht. 2023.
\newblock \href {https://ideas.repec.org/p/zbw/i4rdps/43.html} {{Replication Report on Altmann et al. (2022)}}.
\newblock I4R Discussion Paper Series~43, The Institute for Replication (I4R).

\bibitem[{Begley and Ioannidis(2015)}]{begley2015reproducibility}
C.~Glenn Begley and John P.~A. Ioannidis. 2015.
\newblock \href {https://doi.org/10.1161/CIRCRESAHA.114.303819} {Reproducibility in science: improving the standard for basic and preclinical research}.
\newblock \emph{Circulation Research}, 116(1):116--126.

\bibitem[{Brodeur et~al.(2024)Brodeur, Mikola, Cook, Brailey, Briggs, de~Gendre, Dupraz, Fiala, Gabani et~al.}]{RePEc:zbw:i4rdps:107}
Abel Brodeur, Derek Mikola, Nikolai Cook, Thomas Brailey, Ryan Briggs, Alexandra de~Gendre, Yannick Dupraz, Lenka Fiala, Jacopo Gabani, et~al. 2024.
\newblock \href {https://EconPapers.repec.org/RePEc:zbw:i4rdps:107} {Mass reproducibility and replicability: A new hope}.
\newblock I4R Discussion Paper Series 107, The Institute for Replication (I4R).

\bibitem[{Camerer et~al.(2018)Camerer, Dreber, Holzmeister, Ho, Huber, Johannesson, Kirchler, Nave, Nosek et~al.}]{camerer2018evaluating}
Colin~F Camerer, Anna Dreber, Felix Holzmeister, Teck-Hua Ho, J{\"u}rgen Huber, Magnus Johannesson, Michael Kirchler, Gideon Nave, Brian~A Nosek, et~al. 2018.
\newblock Evaluating the replicability of social science experiments in nature and science between 2010 and 2015.
\newblock \emph{Nature human behaviour}, 2(9):637--644.

\bibitem[{Chen et~al.(2024)Chen, Gangji, Karim, McCanny, and Webb}]{chen2024many}
Shi Chen, Areez Gangji, Sunny Karim, Anthony McCanny, and Matthew~D. Webb. 2024.
\newblock The many misspellings of albuquerque: A comment on 'sorting or steering: The effects of housing discrimination on neighborhood choice'.
\newblock I4R Discussion Paper Series 108, The Institute for Replication (I4R).

\bibitem[{Christensen and Timmins(2018)}]{NBERw24826}
Peter Christensen and Christopher Timmins. 2018.
\newblock \href {https://doi.org/10.3386/w24826} {Sorting or steering: The effects of housing discrimination on neighborhood choice}.
\newblock Working Paper 24826, National Bureau of Economic Research.

\bibitem[{Collaboration(2012)}]{OpenScienceCollaboration2012}
Open~Science Collaboration. 2012.
\newblock \href {https://doi.org/10.1177/1745691612462588} {An open, large-scale, collaborative effort to estimate the reproducibility of psychological science}.
\newblock \emph{Perspectives on Psychological Science}, 7(6):657--660.

\bibitem[{Collaboration(2015)}]{reproducibility_project2015}
Open~Science Collaboration. 2015.
\newblock \href {https://doi.org/10.1126/science.aac4716} {Estimating the reproducibility of psychological science}.
\newblock \emph{Science}, 349(6251):aac4716.

\bibitem[{Collaboration(2016)}]{project_wiki}
Open~Science Collaboration. 2016.
\newblock \href {https://osf.io/ezcuj/wiki/home/} {[link]}.

\bibitem[{Daarstad et~al.(2023)Daarstad, Park, and Balogh}]{RePEc:zbw:i4rdps:97}
Haley Daarstad, RyuGyung Park, and Timea Balogh. 2023.
\newblock \href {https://ideas.repec.org/p/zbw/i4rdps/97.html} {{A comment on Herzog, Baron, and Gibbons (2022)}}.
\newblock I4R Discussion Paper Series~97, The Institute for Replication (I4R).

\bibitem[{Davey(2022)}]{Davey2022}
Reginald Davey. 2022.
\newblock \href {https://www.news-medical.net/life-sciences/What-is-the-Replication-Crisis.aspx} {What is the replication crisis?}
\newblock \emph{News-Medical}.
\newblock Retrieved on October 10, 2024.

\bibitem[{Gravitas(2023)}]{AutoGPT}
Significant Gravitas. 2023.
\newblock \href {https://github.com/Significant-Gravitas/AutoGPT} {Auto-gpt: An autonomous gpt-4 experiment}.
\newblock GitHub repository.

\bibitem[{Gsottbauer et~al.(2022)Gsottbauer, Müller, Müller, Trautmann, and Zudenkova}]{RePEc:oup:econjl:v:132:y:2022:i:647:p:2392-2411.}
Elisabeth Gsottbauer, Daniel Müller, Samuel Müller, Stefan~T Trautmann, and Galina Zudenkova. 2022.
\newblock \href {https://ideas.repec.org/a/oup/econjl/v132y2022i647p2392-2411..html} {{Social Class and (Un)Ethical Behaviour: Causal and Correlational Evidence}}.
\newblock \emph{The Economic Journal}, 132(647):2392--2411.

\bibitem[{Herzog et~al.(2022)Herzog, Baron, and Gibbons}]{doi:10.1086/714924}
Stephen Herzog, Jonathon Baron, and Rebecca~Davis Gibbons. 2022.
\newblock \href {https://doi.org/10.1086/714924} {Antinormative messaging, group cues, and the nuclear ban treaty}.
\newblock \emph{The Journal of Politics}, 84(1):591--596.

\bibitem[{Hu et~al.(2024)Hu, Peters, and Kang}]{hu2025leap}
Chuxuan Hu, Austin Peters, and Daniel Kang. 2024.
\newblock \href {https://doi.org/10.14778/3705829.3705843} {Leap: Llm-powered end-to-end automatic library for processing social science queries on unstructured data}.
\newblock \emph{Proc. VLDB Endow.}, 18(2):253–264.

\bibitem[{Jimenez et~al.(2024)Jimenez, Yang, Wettig, Yao, Pei, Press, and Narasimhan}]{jimenez2024swebenchlanguagemodelsresolve}
Carlos~E. Jimenez, John Yang, Alexander Wettig, Shunyu Yao, Kexin Pei, Ofir Press, and Karthik Narasimhan. 2024.
\newblock \href {https://arxiv.org/abs/2310.06770} {Swe-bench: Can language models resolve real-world github issues?}
\newblock \emph{Preprint}, arXiv:2310.06770.

\bibitem[{Kjelsrud et~al.(2023)Kjelsrud, Kotsadam, and Rogeberg}]{paper1-report}
Anders Kjelsrud, Andreas Kotsadam, and Ole Rogeberg. 2023.
\newblock \href {https://ideas.repec.org/p/zbw/i4rdps/20.html} {{Cooperative Property Rights and Development: Evidence from Land Reform in El Salvador: A Comment}}.
\newblock I4R Discussion Paper Series~20, The Institute for Replication (I4R).

\bibitem[{Montero(2022)}]{paper1}
Eduardo Montero. 2022.
\newblock \href {https://doi.org/10.1086/717042} {Cooperative property rights and development: Evidence from land reform in el salvador}.
\newblock \emph{Journal of Political Economy}, 130(1):48--93.

\bibitem[{{National Academies of Sciences, Engineering, and Medicine}(2019)}]{NAP25303}
{National Academies of Sciences, Engineering, and Medicine}. 2019.
\newblock \href {https://doi.org/10.17226/25303} {\emph{Reproducibility and Replicability in Science}}.
\newblock The National Academies Press, Washington, DC.

\bibitem[{{National Science Foundation}(2022)}]{nsf2023}
{National Science Foundation}. 2022.
\newblock \href {https://www.nsf.gov/pubs/2023/nsf23018/nsf23018.jsp} {Dear colleague letter: Reproducibility and replicability in science}.
\newblock Accessed: 2024-10-12.

\bibitem[{Ono and Zilis(2022)}]{paper2}
Yoshikuni Ono and Michael~A. Zilis. 2022.
\newblock \href {https://doi.org/10.1111/ajps.12599} {Ascriptive characteristics and perceptions of impropriety in the rule of law: Race, gender, and public assessments of whether judges can be impartial}.
\newblock \emph{American Journal of Political Science}, 66(1):43--58.

\bibitem[{OpenAI(2024)}]{gpt4o}
OpenAI. 2024.
\newblock \href {https://platform.openai.com/docs/models/gpt-4o} {[link]}.

\bibitem[{{RePEc}(2024)}]{RePEcTopCitations}
{RePEc}. 2024.
\newblock Top economics publications by number of citations.
\newblock \url{https://ideas.repec.org/top/top.journals.all.html}.
\newblock IDEAS, Federal Reserve Bank of St. Louis.

\bibitem[{Siegel et~al.(2024)Siegel, Kapoor, Nagdir, Stroebl, and Narayanan}]{siegel2024corebenchfosteringcredibilitypublished}
Zachary~S. Siegel, Sayash Kapoor, Nitya Nagdir, Benedikt Stroebl, and Arvind Narayanan. 2024.
\newblock \href {https://arxiv.org/abs/2409.11363} {Core-bench: Fostering the credibility of published research through a computational reproducibility agent benchmark}.
\newblock \emph{Preprint}, arXiv:2409.11363.

\bibitem[{Spearman(1904)}]{spearman1904}
C.~Spearman. 1904.
\newblock \href {http://www.jstor.org/stable/1412159} {The proof and measurement of association between two things}.
\newblock \emph{The American Journal of Psychology}, 15(1):72--101.

\bibitem[{SSRP(2024)}]{ssrp}
SSRP. 2024.
\newblock \href {https://www.socialsciencereproduction.org/metrics} {[link]}.

\bibitem[{Sun et~al.(2024)Sun, Zheng, Xie, Liu, Chu, Qiu, Xu, Ding, Li, Geng, Wu, Wang, Chen, Yin, Ren, Fu, He, Yuan, Liu, Liu, Li, Dong, Cheng, Zhang, Heng, Dai, Luo, Wang, Wen, Qiu, Guo, Xiong, Liu, and Li}]{sun2024surveyreasoningfoundationmodels}
Jiankai Sun, Chuanyang Zheng, Enze Xie, Zhengying Liu, Ruihang Chu, Jianing Qiu, Jiaqi Xu, Mingyu Ding, Hongyang Li, Mengzhe Geng, Yue Wu, Wenhai Wang, Junsong Chen, Zhangyue Yin, Xiaozhe Ren, Jie Fu, Junxian He, Wu~Yuan, Qi~Liu, Xihui Liu, Yu~Li, Hao Dong, Yu~Cheng, Ming Zhang, Pheng~Ann Heng, Jifeng Dai, Ping Luo, Jingdong Wang, Ji-Rong Wen, Xipeng Qiu, Yike Guo, Hui Xiong, Qun Liu, and Zhenguo Li. 2024.
\newblock \href {https://arxiv.org/abs/2312.11562} {A survey of reasoning with foundation models}.
\newblock \emph{Preprint}, arXiv:2312.11562.

\bibitem[{Tian et~al.(2024)Tian, Gao, Zhang, Chen, Fan, Guo, Haas, Ji, Krongchon et~al.}]{tian2024scicoderesearchcodingbenchmark}
Minyang Tian, Luyu Gao, Shizhuo~Dylan Zhang, Xinan Chen, Cunwei Fan, Xuefei Guo, Roland Haas, Pan Ji, Kittithat Krongchon, et~al. 2024.
\newblock \href {https://arxiv.org/abs/2407.13168} {Scicode: A research coding benchmark curated by scientists}.
\newblock \emph{Preprint}, arXiv:2407.13168.

\bibitem[{Yang et~al.(2024)Yang, Jimenez, Wettig, Lieret, Yao, Narasimhan, and Press}]{yang2024sweagent}
John Yang, Carlos~E. Jimenez, Alexander Wettig, Kilian Lieret, Shunyu Yao, Karthik Narasimhan, and Ofir Press. 2024.
\newblock \href {https://arxiv.org/abs/2405.15793} {Swe-agent: Agent-computer interfaces enable automated software engineering}.
\newblock \emph{Preprint}, arXiv:2405.15793.

\bibitem[{Yao et~al.(2023)Yao, Chen, Yang, and Narasimhan}]{yao2023webshopscalablerealworldweb}
Shunyu Yao, Howard Chen, John Yang, and Karthik Narasimhan. 2023.
\newblock \href {https://arxiv.org/abs/2207.01206} {Webshop: Towards scalable real-world web interaction with grounded language agents}.
\newblock \emph{Preprint}, arXiv:2207.01206.

\end{thebibliography}




\end{document}